\documentclass[10pt,twocolumn,letterpaper]{article}

\usepackage{cvpr}              % To produce the CAMERA-READY version
\usepackage{graphicx}
\usepackage{multirow}

\usepackage[dvipsnames]{xcolor}

\definecolor{cvprblue}{rgb}{0.21,0.49,0.74}
\usepackage[pagebackref,breaklinks,colorlinks,citecolor=cvprblue]{hyperref}

\def\paperID{2093} % *** Enter the Paper ID here
\def\confName{CVPR}
\def\confYear{2024}

\title{HMD-Poser: On-Device Real-time Human Motion Tracking from Scalable Sparse Observations}

\author{Peng Dai, Yang Zhang, Tao Liu, Zhen Fan, Tianyuan Du, \\
Zhuo Su, Xiaozheng Zheng, Zeming Li, \\
PICO, ByteDance \\
{\tt\small \{daipeng.2022, zhangyang.0621, liutao.96, fanzhen.0315, dutianyuan, } \\
{\tt\small suzhuo, zhengxiaozheng, lizeming.001\}@bytedance.com}
}

\begin{document}
\maketitle

% \input{sec/0_abstract}    
% \input{sec/1_intro}
% \input{sec/2_formatting}
% \input{sec/3_finalcopy}

%%%%%%%%% ABSTRACT
\begin{abstract}
% Real-time human motion tracking is in high demand for various VR applications. Nevertheless, it is especially challenging to implement on a standalone VR Head-Mounted Display (HMD) such as Meta Quest and PICO. 
It is especially challenging to achieve real-time human motion tracking on a standalone VR Head-Mounted Display (HMD) such as Meta Quest and PICO.
In this paper, we propose HMD-Poser, the first unified approach to recover full-body motions using scalable sparse observations from HMD and body-worn IMUs. In particular, it can support a variety of input scenarios, such as HMD, HMD+2IMUs, HMD+3IMUs, etc. The scalability of inputs may accommodate users' choices for both high tracking accuracy and easy-to-wear. A lightweight temporal-spatial feature learning network is proposed in HMD-Poser to guarantee that the model runs in real-time on HMDs. Furthermore, HMD-Poser presents online body shape estimation to improve the position accuracy of body joints. Extensive experimental results on the challenging AMASS dataset show that HMD-Poser achieves new state-of-the-art results in both accuracy and real-time performance. We also build a new free-dancing motion dataset to evaluate HMD-Poser's on-device performance and investigate the performance gap between synthetic data and real-captured sensor data. Finally, we demonstrate our HMD-Poser with a real-time Avatar-driving application on a commercial HMD. Our code and free-dancing motion dataset are available \href{https://pico-ai-team.github.io/hmd-poser}{here}.
\end{abstract}

%%%%%%%%% Introduction
\section{Introduction}
\label{sec:intro}
Human motion tracking (HMT), which aims at estimating the orientations and positions of body joints in 3D space, is highly demanded in various VR applications, such as gaming and social interaction. However, it is quite challenging to achieve both accurate and real-time HMT on HMDs. There are two main reasons. First, since only the user's head and hands are tracked by HMD (including hand controllers) in the typical VR setting, estimating the user's full-body motions, especially lower-body motions, is inherently an under-constrained problem with such sparse tracking signals. Second, computing resources are usually highly restricted in portable HMDs, which makes deploying a real-time HMT model on HMDs even harder.

\begin{figure}
    \centering
    \includegraphics[width=0.95\linewidth]{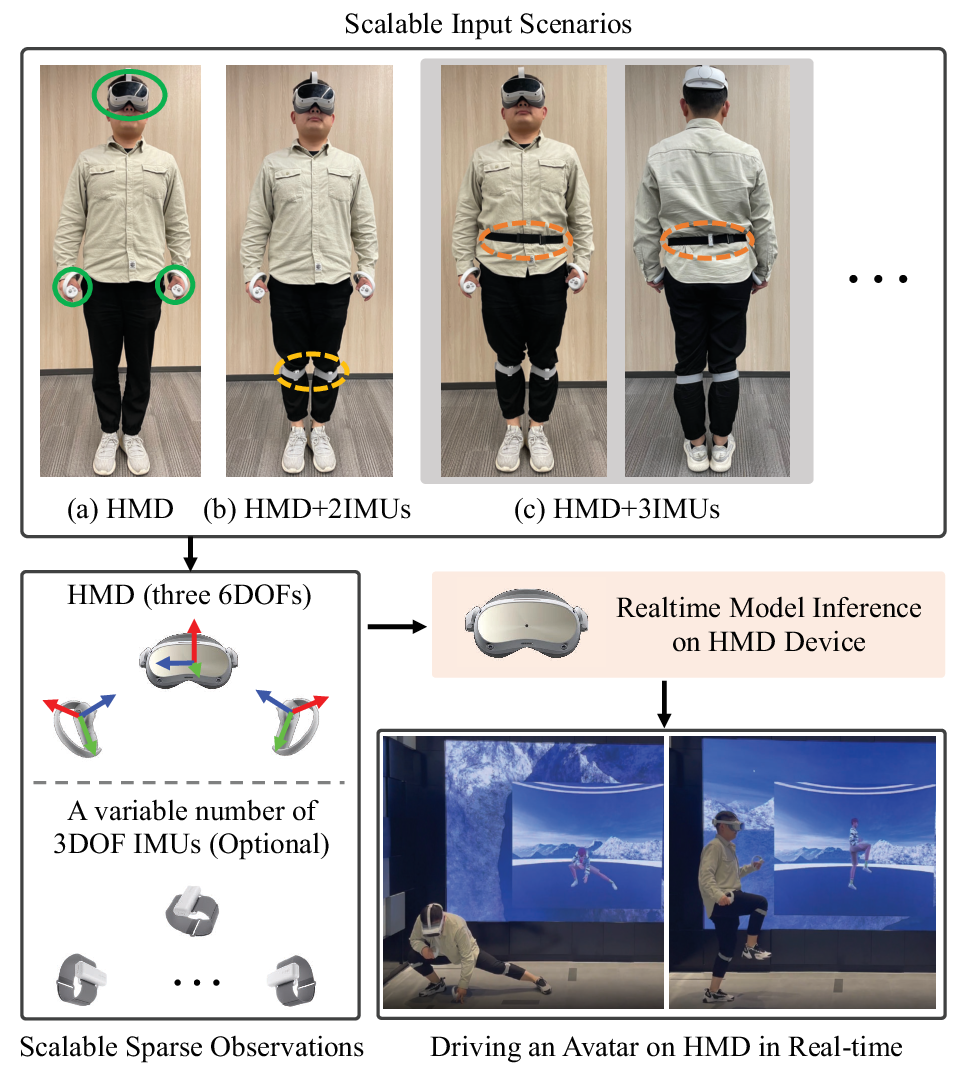}
    \caption{HMD-Poser can handle scalable input scenarios, including (a) HMD, (b) HMD+2IMUs wherein two IMUs are worn on the lower legs, (c) HMD+3IMUs wherein a third IMU is added to the pelvis, etc. HMD-Poser runs on HMD and outputs full-body motion data to drive an Avatar in real-time.}
    \label{fig:inputconfigure}
\end{figure}

Prior works have focused on improving the accuracy of full-body tracking. One category of methods utilizes three 6DOFs (degrees of freedom) from HMD to estimate full-body motions, and they could be roughly classified into the physics-simulator-driven methods~\cite{lee2023questenvsim, winkler2022questsim} and the data-driven methods~\cite{aliakbarian2022flag, dittadi2021full, du2023avatars, aliakbarian2023hmdnemo, jiang2022avatarposer, zheng2023realistic}. These methods usually have difficulties in some uncorrelated upper-lower body motions where different lower-body movements are represented by similar upper-body observations. As a result, it's hard for them to accurately drive an Avatar with unlimited movements in VR applications. The other category of methods~\cite{huang2018deep, yi2021transpose, yi2022pip, yi2023egolocate, pan2023fusing} uses six 3DOF IMUs (inertial measurement units) worn on the user's head, forearms, pelvis, and lower legs respectively for HMT. While these methods could improve lower-body tracking accuracy by adding legs' IMU data, it's theoretically difficult for them to provide accurate body joint positions due to the inherent drifting problem of IMU sensors. 
Recently, SparsePoser~\cite{ponton2023sparseposer} combined HMD with three 6DOF trackers on the pelvis and feet to improve accuracy. However, 6DOF trackers usually need extra base stations which make them user-unfriendly and they are much more expensive than 3DOF IMUs.

Different from existing methods, we propose HMD-Poser to combine HMD with scalable 3DOF IMUs. Considering users' preferences between easy-to-wear and high accuracy, HMD-Poser designs a unified framework to be compatible with scalable observations, as shown in Fig.~\ref{fig:inputconfigure}. Scalability means it can handle multiple input scenarios, including a) HMD, b) HMD+2IMUs, c) HMD+3IMUs, etc. 
Furthermore, unlike existing works that use the same default shape parameters for joint position calculation, our HMD-Poser involves hand representations relative to the head coordinate frame to estimate the user's body shape parameters online. It can improve the joint position accuracy when the users' body shapes vary in real applications.
% In this paper, we address the above issues through HMD-Poser, a novel deep learning-based method that combines 6DOF HMD data with 3DOF IMU data for real-time HMT. Considering users' preferences between easy-to-wear and high accuracy, HMD-Poser designs a unified framework to be compatible with scalable sparse observations, as shown in Fig.~\ref{fig:inputconfigure}. Scalability means it can handle multiple input scenarios, including (a) HMD, (b) HMD+2IMUs, (c) HMD+3IMUs, etc. 
% Existing works tend to use the same default shape parameters for joint position calculation. However, it is impossible to obtain accurate joint positions when the users' body shapes vary in real applications.

Real-time on-device execution is another key factor that affects users' VR experience. Nevertheless, it has been overlooked in most existing methods. Recent methods~\cite{du2023avatars, jiang2022avatarposer, zheng2023realistic} usually adopt the clip setting, i.e., processing all input data within a clip during each model inference, which may increase computational cost and time delay. 
% HMD-NeMo~\cite{aliakbarian2023hmdnemo} uses gated recurrent units (GRUs)~\cite{cho2014properties} and Transformer~\cite{vaswani2017attention} networks to achieve the potential of real-time HMT, but it did not demonstrate the on-device deployment or online Avatar animations in their paper. 
Motivated by HMD-NeMo~\cite{aliakbarian2023hmdnemo}, our HMD-Poser introduces a lightweight temporal-spatial feature learning (TSFL) network that combines long short-term memory (LSTM)~\cite{hochreiter1997long} networks for temporal feature capturing with Transformer~\cite{vaswani2017attention} encoders for spatial correlation learning. With the help of the hidden state in LSTM, the input length and computational cost of the Transformer are significantly reduced, making the model real-time runnable on HMDs.

Our contributions are concluded as follows: 
(1) To the best of our knowledge, HMD-Poser is the first HMT solution that designs a unified framework to handle scalable sparse observations from HMD and wearable IMUs. Hence, it could recover accurate full-body poses with fewer positional drifts. 
(2) HMD-Poser builds a simple yet effective network by combining a set of standard components, such as LSTM~\cite{hochreiter1997long}, Transformer~\cite{vaswani2017attention}, etc. It achieves state-of-the-art results on the AMASS dataset and runs in real-time on consumer-grade HMDs. 
% HMD-Poser proposes a lightweight TSFL network combining LSTM with Transformer for feature learning and introduces online body shape estimation for joint position calculation, making it achieve state-of-the-art results on the AMASS dataset and run in real-time on HMDs. 
(3) A free-dancing motion capture dataset is built for on-device evaluation. It is the first dataset that contains synchronized ground-truth 3D human motions and real-captured HMD and IMU sensor data.

\begin{figure*}[!ht]
    \centering
    \includegraphics[width=0.95\linewidth]{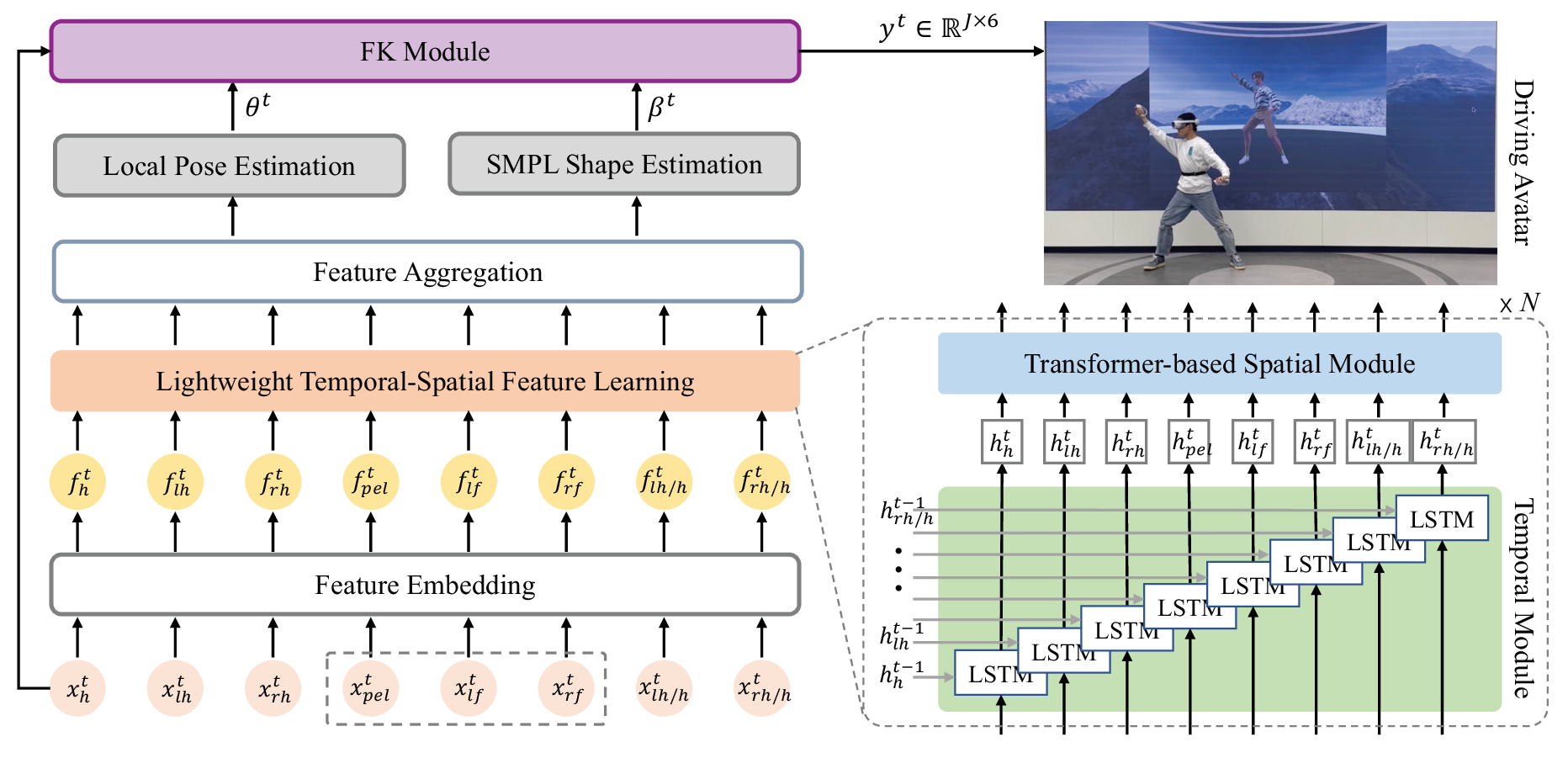}
    \caption{Overview of HMD-Poser. At each time step $t$, each component in the input data $x^t$ (see Eq.~\ref{eq:input_feat}) is firstly mapped to a higher-dimensional embedding feature $f^t$ via the feature embedding module. Then, a lightweight temporal-spatial feature learning network is adopted to generate representations with rich temporal and spatial correlation information. Next, two regression heads regress the local pose parameters $\theta^t$ and the shape parameters $\beta^t$ of SMPL, respectively. Finally, a forward-kinematics (FK) module is adopted to calculate the global poses and positions of all joints which are used to drive an Avatar in real-time.}
    \label{fig:framework}
\end{figure*}

%%%%%%%%% Related Work
\section{Related Work}
\label{sec:relatedwork}
% HMT has attracted much interest in recent years since it can be applied to various applications. 
HMT has attracted much interest in recent years. Existing works generate tracking results from optical markers~\cite{zanfir2021thundr, ghorbani2021soma, loper2014mosh}, depth sensors~\cite{UnstructureLan, robustfusion, bashirov2021real, su2022robustfusion, yu2018doublefusion}, monocular images~\cite{bogo2016keep, HMR18, joo2020eft, SMPL-X:2019, Kanazawa_2019_CVPR, Kocabas_2020_CVPR, Kocabas_2021_ICCV, tripathi20233d, huang2022neural, li2023niki}, ego-centric views~\cite{akada2022unrealego, wang2023scene, wang2021estimating, li2023ego}, single-view videos~\cite{zheng20213d,chen2022learning, li2022mhformer, zhang2022mixste, zhan2022ray3d, gartner2022trajectory}, and multi-view videos~\cite{li20213d, zhang2022voxeltrack, mitra2020multiview}. Recently, methods using sparse signals from HMD or wearable IMU sensors, have received more attention~\cite{yi2021transpose, yi2022pip, jiang2022avatarposer, aliakbarian2022flag, du2023avatars}.

\subsection{HMT in HMD Setting}
\label{sec:relatedwork_HMD}
In a typical VR HMD setting, the upper body is tracked by signals from HMD with hand controllers, while the lower body's tracking signals are absent. One advantage of this setting is that HMD could provide reliable global positions of the user's head and hands with SLAM, rather than only 3DOF data from IMUs. Existing methods fall into two categories. First is the physics-simulator-based methods~\cite{lee2023questenvsim, winkler2022questsim}. QuestSim~\cite{winkler2022questsim} and QuestEnvSim~\cite{lee2023questenvsim} utilized Nvidia's $IsaacGym$~\cite{makoviychuk2021isaac} for physics simulation and reinforcement learning for model training. However, physics simulators are typically non-differential black boxes, making these methods incompatible with existing machine learning frameworks and difficult to deploy to HMDs. Second is the data-driven methods~\cite{aliakbarian2022flag, dittadi2021full, du2023avatars, aliakbarian2023hmdnemo, jiang2022avatarposer, zheng2023realistic}. 
HMD-NeMo~\cite{aliakbarian2023hmdnemo}, Avatarposer~\cite{jiang2022avatarposer}, and AvatarJLM~\cite{zheng2023realistic} relied on regression models to establish a direct mapping from sparse tracking data to target 3D human motions. Other works tried to improve performance by generative models, such as normalizing flows~\cite{aliakbarian2022flag}, Variational Autoencoders (VAE)~\cite{dittadi2021full} and diffusion models~\cite{du2023avatars}. Although these methods have shown promising results on public datasets such as AMASS~\cite{mahmood2019amass}, they still have many failure cases, e.g., when the upper body remains stationary but the lower legs move.

\subsection{HMT from Wearable IMUs}
\label{sec:relatedwork_imus}
Recent studies~\cite{von2017sparse,huang2018deep,yi2021transpose,yi2022pip,jiang2022transformer} have explored the use of six IMUs, which track the signals of the user's head, fore-arms, lower-legs, and pelvis respectively, for full-body motion estimation. The pioneering work is the optimization-based approach SIP~\cite{von2017sparse}, which demonstrated the feasibility of reconstructing accurate 3D full-body motion by only six IMUs, albeit with limited speed. Subsequently, a few deep learning-based approaches~\cite{huang2018deep, nagaraj2020rnn} used recurrent neural networks (RNN) to improve real-time performance but were unable to estimate the global translation. TransPose~\cite{yi2021transpose} tried to estimate the global translation by fusing a supporting-foot-based method with an RNN-based root translation regression model. Moreover, PIP~\cite{yi2022pip} introduced a physics-aware motion optimizer and TIP~\cite{jiang2022transformer} proposed a stationary-body-points predictor to further improve the translation accuracy and physical plausibility. However, these methods are prone to positional drift due to the inevitable accumulation errors of IMU sensors, making it difficult to provide accurate joint positions.

% To obtain accurate full-body orientations and joint positions at the same time, our proposed HMD-Poser combines the HMD setting with scalable IMUs. IMU sensors worn on the pelvis and lower legs help to improve the lower body pose estimation, and full-body joint positions are derived from the reliable head position from HMD.
HMD-Poser combines the HMD setting with scalable IMUs. IMUs worn on the pelvis and lower legs improve the lower body pose estimation, and full-body joint positions are derived from the reliable head position from HMD.

%%%%%%%%% Proposed Method
\section{Method}
\label{sec:method}
\subsection{Overview}
\label{sec:problemdefine}
\noindent
\textbf{Task.} 
Our task is to estimate both the orientations and positions of all body joints in real-time using HMD and a variable number of IMUs. Specifically, we consider three input scenarios as shown in Fig.~\ref{fig:inputconfigure}. Intuitively, the HMD scenario is the most convenient for users among the three scenarios, but with the worst tracking accuracy. While the HMD+3IMUs scenario is exactly the opposite. The HMD+2IMUs scenario balances user-friendliness and tracking accuracy, making it suitable for most applications. It's worth noting that our method can also be applied to other HMD/IMU configurations.

\noindent
\textbf{Framework.}
As illustrated in Fig.~\ref{fig:framework}, HMD-Poser's pipeline consists of four components. (1) A feature embedding module maps the input data $x^t$ to a higher-dimensional embedding space which serves as the input to subsequent networks. (2) A lightweight TSFL network learns how each input component is related to each other and evolves over time, generating representations with rich temporal and spatial correlation. (3) With the feature aggregated from TSFL, Two MLP-based (multi-layer perceptron) heads regress the local pose parameters $\theta^t$ and the shape parameters $\beta^t$ of SMPL, respectively. (4) Given $\theta^t$, $\beta^t$ and the head position provided by HMD, a forward-kinematics (FK) module calculates all joint positions and concatenates them with $\theta^t$ to generate the final output $y^t\in \mathbb R ^{(J\times 6)}$. Following previous works~\cite{jiang2022avatarposer,yi2021transpose}, we adopt the first 22 joints of the SMPL~\cite{loper2015smpl} model, i.e., $J=22$.

\subsection{Scalable Input Processing}
\label{sec:scalable_input_processing}
\noindent
\textbf{Input representation.}
The input signal $x^t$ contains HMD's head and hand 6DOF data and optionally the rotation and acceleration measurements from IMUs. We follow AvatarPoser~\cite{jiang2022avatarposer} to use a concatenated vector of position, linear velocity, rotation, and angular velocity to obtain the representation for the head $x^{t}_{h}$, the left hand $x^{t}_{lh}$, and the right hand $x^{t}_{rh}$. As for IMU data, we adopt a concatenated vector of rotation, angular velocity, and acceleration to obtain the representation for the pelvis $x^{t}_{pel}$, the left leg $x^{t}_{lf}$, and the right leg $x^{t}_{rf}$. We represent rotation and angular velocity with a 6D vector due to its simplicity and continuity~\cite{zhou2019continuity}, resulting in $x^{t}_{h}, x^{t}_{lh}, x^{t}_{rh}\in \mathbb R^{1\times 18}$ and $x^{t}_{pel}, x^{t}_{lf}, x^{t}_{rf}\in \mathbb R^{1\times 15}$. All IMU rotation and acceleration data are calibrated to a common body-centric frame before feeding to the model. To better estimate the shape parameters and improve the joint position accuracy (see Sec.~\ref{sec:experiment} for more experimental results), we additionally calculate the hand representations relative to the head coordinate frame, $x^{t}_{lh/h} \in \mathbb R^{1\times 18}$ and $x^{t}_{rh/h} \in \mathbb R^{1\times 18}$.
Overall, the input data, $x^t \in \mathbb R^{1\times 135}$, can be written as
\begin{equation}
\label{eq:input_feat} 
x^t=[x^{t}_{h}, x^{t}_{lh}, x^{t}_{rh}, x^{t}_{pel}, x^{t}_{lf}, x^{t}_{rf}, x^{t}_{lh/h}, x^{t}_{rh/h} ]
\end{equation}

As described in Sec.~\ref{sec:problemdefine}, our model needs to be compatible with scalable inputs. It means that the IMU data of the pelvis $x^{t}_{pel}$ is missing in the second scenario, and all the IMU data ($x^{t}_{pel}$, $x^{t}_{lf}$ and $x^{t}_{rf}$) is missing in the first scenario. To make our model compatible with all scenarios in a unified framework, the feature dimension of the input data remains the same for all scenarios, and the missing observations are filled with zero-padding. 
% A corresponding mask is used in the subsequent Transformer to avoid computing self-attention on the padded values.

\noindent
\textbf{Feature embedding.} 
A set of fully connected (FC) layers is adopted to project raw input data to a higher-dimensional embedding space. Note that eight components within the input stream, as shown in Eq.~\ref{eq:input_feat}, are processed independently, then generating the embedding representations $[f^{t}_{h}, f^{t}_{lh}, f^{t}_{rh}, f^{t}_{pel}, f^{t}_{lf}, f^{t}_{rf}, f^{t}_{lh/h}, f^{t}_{rh/h}]$. For each component, since the range of values corresponding to the orientations is different from those of the positions, we follow~\cite{aliakbarian2023hmdnemo} to decouple such information and embed them via separate FC layers and concatenate the results back together.

\subsection{Lightweight TSFL Network}
\label{sec:leightweightTSFL}
After the embedding layer, components in the input are still temporally isolated and spatially independent of each other. In other words, it lacks temporal and spatial correlation information, which is the key to tracking accurate human motions. To solve this problem, two representative models, i.e., Transformer and RNN, are adopted for temporal and spatial feature learning in existing methods. Although Transformer-based methods~\cite{jiang2022avatarposer, zheng2023realistic} have achieved state-of-the-art results in HMT, their computational costs are much higher than RNN-based methods~\cite{yi2021transpose, yi2022pip} as Transformer does not preserve the hidden state and needs to recompute the entire history in the video clip at each time step. 
Therefore, the current Transformer-based methods are not suitable for HMD deployment.

For a sequence of length $M$, the time complexity of a standard Transformer block is $\mathcal{O}(M^{2}d+Md^2)$ where $d$ is the dimension of the hidden state. It means that the Transformer has a quadratic time complexity with respect to the sequence length in attention layers.
To achieve both accurate and real-time human motion tracking, HMD-Poser introduces a lightweight TSFL network that combines the RNN model with the Transformer. As shown in Fig.~\ref{fig:framework}, the lightweight TSFL network is composed of a stack of $N=2$ identical blocks. And each block has two sub-blocks. The first is a set of long short-term memory (LSTM) modules to independently learn the temporal representation of each component in the input, and the second is a Transformer encoder to learn the spatial correlation among different components. 
The time complexity of the LSTM model is $\mathcal{O}(d^2)$, which is negligible compared to that of the Transformer model.
With the help of the hidden state in LSTM, the Transformer could focus on spatial feature learning within each frame.
It means that the sequential length $M$ in our Transformer is reduced to the number of input components, i.e., $M=8$, which is much smaller than that in previous methods (e.g., $M=40$ in~\cite{jiang2022avatarposer} and $M=45$ in~\cite{zheng2023realistic}). As a result, our method is more than 5 times faster than previous Transformer-based methods~\cite{jiang2022avatarposer, zheng2023realistic} in terms of a single Transformer layer.
Meanwhile, by introducing LSTM to retain complete historical information, our TSFL network can result in accuracy improvements for long-period motions.

\subsection{Position Estimation with Shape Head}
\label{sec:positionestimation_withshape}
\noindent
\textbf{Pose and shape estimation.} 
Most previous methods~\cite{aliakbarian2023hmdnemo, du2023avatars, jiang2022avatarposer, zheng2023realistic} only considered the pose parameters and ignored the shape parameters. In other words, they used the same default shape parameters to calculate joint positions. We argue that these methods are not optimal in practical applications, because the shape parameters usually vary by different users. This would lead to problems such as penetration, skating, and joint position errors, especially when the difference between the user's shape parameters and the default is significant. To solve this problem, HMD-Poser adopts two regression heads named pose head and shape head. As shown in Fig.~\ref{fig:framework}, the pose head aims at regressing the local pose parameters $\theta^t$ of SMPL and the shape head is responsible for shape parameters $\beta^t$ of SMPL. Both regression heads are designed as a 2-layer MLP.

\noindent
\textbf{Forward-Kinematics.} 
The FK module calculates all joint positions given $\theta^t$, $\beta^t$, and the head position in $x^t_h$. We use the differentiable SMPL model~\cite{loper2015smpl}, $\mathcal{M}(\theta, \beta, trans) \in  \mathbb R ^{(6890\times 3)}$, as the FK module. Using the estimated joint positions and their corresponding ground-truth values can (1) train the shape head and (2) assist in reducing the accumulating error of pose estimation along the kinematic chain.
 
\begin{table*}[!htb]
  \centering
  \begin{tabular}{@{}lcccccccc@{}}
    \toprule
    Method & MPJRE$\downarrow$ & MPJPE$\downarrow$ & MPJVE$\downarrow$ & Jitter$\downarrow$ & H-PE$\downarrow$ & U-PE$\downarrow$ & L-PE$\downarrow$ & R-PE$\downarrow$ \\
    \midrule
    \multirow{3}{*}{}AvatarPoser~\cite{jiang2022avatarposer} &  2.94 & 5.84 & 26.60 & 13.97 & 4.58 & 3.24 & 9.59 & 5.05 \\
    AGRoL~\cite{du2023avatars} &  2.70 & 5.73 & 19.08 & 7.65 & 4.29 & 3.16 & 9.44 & 5.15 \\ 
    AvatarJLM~\cite{zheng2023realistic} &  2.81 & 5.03 & 20.91 & 6.94 & 2.01 & 3.00 & 7.96 & 4.58 \\
    \midrule
    \multirow{2}{*}{}Transpose~\cite{yi2021transpose} & 3.05 & 4.57 & 22.41 & 7.98 & 3.83 & 3.05 & 6.76 & 4.62\\
    PIP~\cite{yi2022pip} &  2.45 & 4.54 & 19.02 & 8.13 & 4.54 & 3.15 & 6.53 & 4.54 \\ 
    \midrule
    \multirow{3}{*}{}HMD-Poser: HMD &  2.28 & 3.19 & 17.47 & 6.07 & 1.65 & 1.67 & 5.40 & 3.02 \\
    HMD-Poser: HMD+2IMUs &  1.83 & 2.27 & 13.28 & 5.96 & 1.39 & 1.51 & 3.35 & 2.74 \\
    HMD-Poser: HMD+3IMUs &  \textbf{1.73} & \textbf{1.89} & \textbf{11.03} & \textbf{5.35} & \textbf{1.27} & \textbf{1.46} & \textbf{2.46} & \textbf{2.37} \\
    \bottomrule
  \end{tabular}
  \caption{Comparison with state-of-the-art HMD-based and 6IMUs-based methods on protocol1. We retrain existing approaches with their public source code and training data on this protocol. Note that we also provide head and hand positions to Transpose and PIP for a fair comparison. The best results are in \textbf{bold}.}
  \label{tab:results_protocol1}
\end{table*}

% The FPS is calculated on the NVIDIA GeForce RTX 3080 device. The on-device FPS of our model is presented in Sec.~\ref{sec:liveperformance}. 

\begin{table*}[htb]
  \centering
  \begin{tabular}{@{}lcccccccc@{}}
    \toprule
    Method & MPJRE$\downarrow$ & MPJPE$\downarrow$ & MPJVE$\downarrow$ & Jitter$\downarrow$ & H-PE$\downarrow$ & U-PE$\downarrow$ & L-PE$\downarrow$ & R-PE$\downarrow$ \\
    \midrule
    \multirow{3}{*}{}AvatarPoser~\cite{jiang2022avatarposer} & 4.68 & 6.62 & 33.16 & 10.79 & 3.93 & 2.97 & 11.89 & 5.30 \\
    AGRoL~\cite{du2023avatars} & 4.38 & 6.74 & 24.14 & 6.33 & 3.53 & 3.02 & 12.11 & 5.86 \\
    AvatarJLM~\cite{zheng2023realistic} & 4.45 & 5.96 & 27.50 & 6.91 & 2.30 & 2.97 & 10.28 & 5.22 \\
    \midrule
    \multirow{2}{*}{}Transpose~\cite{yi2021transpose} & 4.31 & 5.29 & 28.18 & 5.16 & 7.38 & 3.86 & 7.36 & 4.80 \\
    PIP~\cite{yi2022pip} & 3.61 & 4.16 & 22.22 & 6.89 & 4.28 & 2.97 & 5.89 & 4.30 \\
    \midrule
    \multirow{3}{*}{}HMD-Poser: HMD & 4.27 & 5.44 & 30.15 & 5.62 & 2.56 & 2.44 & 9.77 & 4.83\\
    HMD-Poser: HMD+2IMUs & 3.66 & 3.68 & 20.29 & 6.22 & \textbf{1.65} & \textbf{2.14} & 5.92 & 4.51 \\
    HMD-Poser: HMD+3IMUs & \textbf{3.49} & \textbf{3.13} & \textbf{16.17} & \textbf{4.93} & 1.81 & 2.17 & \textbf{4.51} & \textbf{3.88} \\
    \bottomrule
  \end{tabular}
  \caption{Comparison to baselines on protocol2. Similarly, we retrain existing approaches with their public source code on this protocol.}
  \label{tab:results_protocol2}
\end{table*}

\subsection{Training HMD-Poser}
\label{sec:modeltraining}
We define the overall loss function $\mathcal{L}$ as a combination of root orientation loss $\mathcal{L}_{ori}$, local pose loss $\mathcal{L}_{lrot}$, global pose loss $\mathcal{L}_{grot}$, joint position loss $\mathcal{L}_{joint}$ and smooth loss $\mathcal{L}_{smooth}$:
\begin{equation}
\label{eq:loss_total} 
\begin{split}
     \mathcal{L} = & \alpha_{ori} \mathcal{L}_{ori} + \alpha_{lrot} \mathcal{L}_{lrot} + \alpha_{grot} \mathcal{L}_{grot} \\
    & + \alpha_{joint} \mathcal{L}_{joint} + \alpha_{smooth} \mathcal{L}_{smooth}
\end{split}
\end{equation}
where $\alpha_{ori}$, $\alpha_{lrot}$, $\alpha_{grot}$, $\alpha_{joint}$, and $\alpha_{smooth}$ are the weights for the respective loss terms. The root orientation loss $\mathcal{L}_{ori}$, local pose loss $\mathcal{L}_{lrot}$, global pose loss $\mathcal{L}_{grot}$, and joint position loss $\mathcal{L}_{joint}$ are calculated as the mean of absolute errors (L1 norm) between the predicted values and the ground-truth values.
To further enhance the temporal smoothness, we define a smooth loss as follows.
\begin{equation}
\label{eq:smoothloss} 
    \mathcal{L}_{smooth} = \frac{1}{(T-2)\times (3J)} \sum_{t=1}^{T-1} \sum_{i=0}^{3J}\left | a^t_i - \hat{a} ^t_i \right |_1  
\end{equation}
where $a^t$ and $\hat{a}^t$ are the computed and the ground-truth acceleration at time $t$, respectively, and $T$ is the sequential length in the training and $J$ is the number of joints.

%%%%%%%%% Experiments
\section{Experiments}
\label{sec:experiment}
In this section, we first compare our method with state-of-the-art methods and conduct ablation studies on the public AMASS~\cite{mahmood2019amass} dataset. Then, we present detailed quantitative and qualitative results on real-captured data using PICO 4 and PICO Motion Trackers. Note that our model can be also deployed to other commercial VR systems that provide the required orientation and position information, such as Meta's Quest2 HMD.

\begin{figure*}[!htb]
    \centering
    \includegraphics[width=1.0\linewidth]{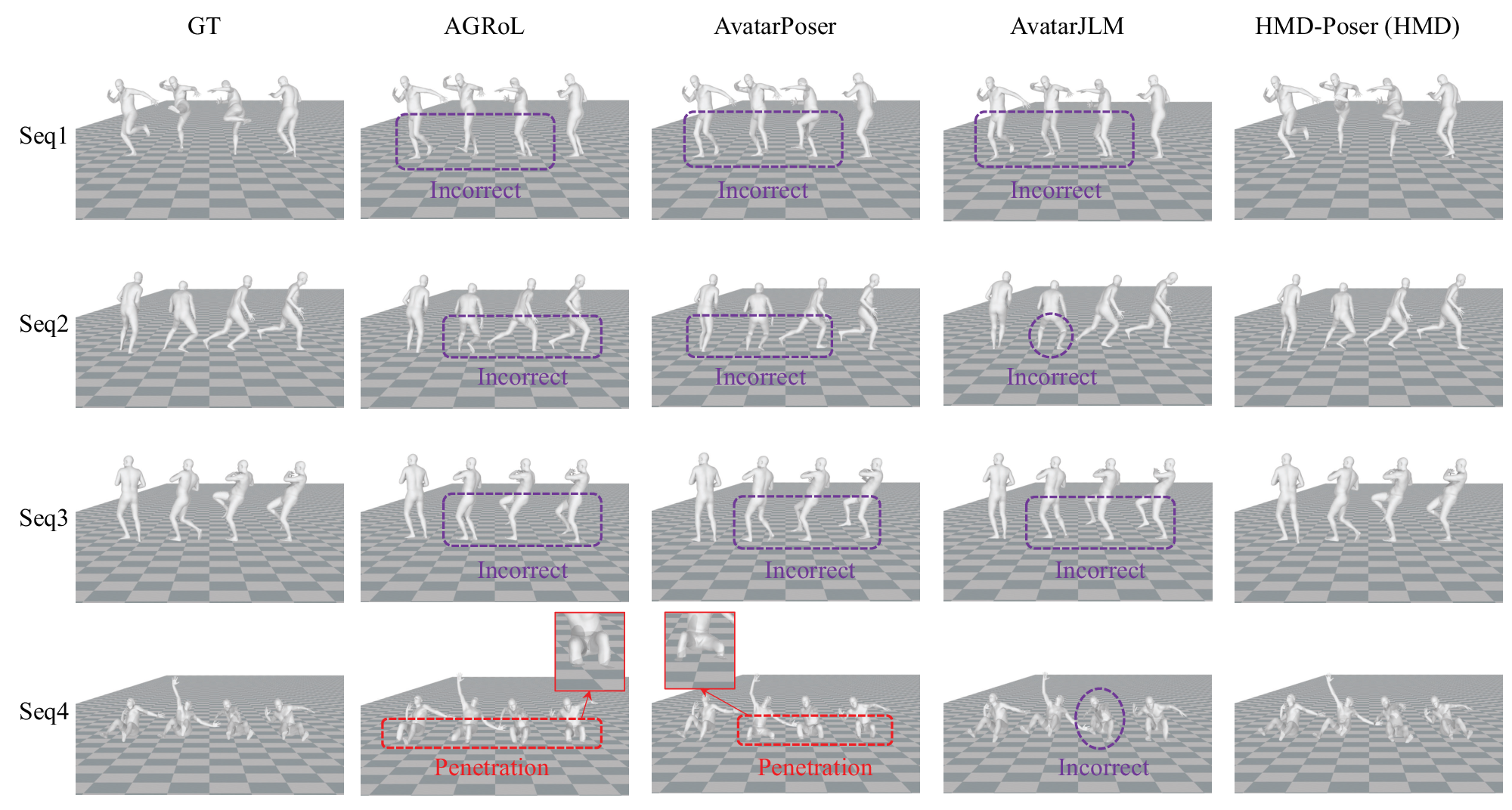}
    \caption{Qualitative comparisons between our method and state-of-the-art methods in HMD setting. When comparing with methods in this category, HMD-Poser uses the HMD input scenario for a fair comparison.}
    \label{fig:qualitative_3IMU}
\end{figure*}

\noindent
\textbf{Implementation details.}
We set $\alpha_{ori}$, $\alpha_{lrot}$, $\alpha_{grot}$, $\alpha_{joint}$, and $\alpha_{smooth}$ in Eq.~\ref{eq:loss_total} to 1.0, 5.0, 1.0, 1.0, and 0.5, respectively. The number of blocks in the TSFL network is set to 2. 
%In each block, a single-layer LSTM and a 3-layer Transformer encoder are used. 
Further details about the model are presented in the supplementary materials. The training data is clipped into short sequences in 40-frame lengths for more effective learning. 
% To optimize the parameters of our HMD-Poser, 
To train HMD-Poser, we use Adam solver~\cite{kingma2014adam} with a batch size of 256. The learning rate starts from $1\times 10^{-3}$ and decays by a factor of 0.1 after 300 epochs. The total number of epochs for training is set to 400.

\noindent
\textbf{Metrics.}
% To evaluate the performance of our method as well as the competing baselines, 
% We use a total of 9 metrics following existing methods~\cite{du2023avatars, jiang2022avatarposer, zheng2023realistic}. 
Following~\cite{du2023avatars}, we use a total of 9 metrics which can be divided into three categories. The first category measures the tracking accuracy and includes the $\textit{MPJRE}$ (Mean Per-Joint Rotation Error [degrees]), $\textit{MPJPE}$ (Mean Per-Joint Position Error [cm]), $\textit{H-PE}$ (Hand), $\textit{U-PE}$ (Upper), $\textit{L-PE}$ (Lower), and $\textit{R-PE}$ (Root). The second category reflects the smoothness of the generated motions and includes the $\textit{MPJVE}$ (Mean Per-Joint Velocity Error [cm/s]) and $\textit{Jitter}$ ($10^2m/s^3$). 
The third category measures the inference speed with the $\textit{FPS}$ (Frames Per Second [Hz]).
% of model inference. We calculate the $\textit{FPS}$ (Frames Per Second [Hz]) of each method.

\subsection{Experiments on AMASS Dataset}
\label{sec:exp_on_amass}
% \subsubsection{Experimental Setup and Implementation Details}
% \label{sec:expsetup}
We follow the recent common practice~\cite{aliakbarian2022flag, du2023avatars, jiang2022avatarposer, zheng2023realistic} of using AMASS~\cite{mahmood2019amass} dataset with two different protocols.
The first protocol uses three subsets CMU~\cite{cmu_amass}, BMLr~\cite{troje2002decomposing} and HDM05~\cite{muller2007mocap} in AMASS, and randomly splits the three datasets into 90$\%$ training data and 10$\%$ testing data. The second protocol includes more subsets in AMASS, using twelve subsets as training data and HumanEva~\cite{sigal2010humaneva} and Transition~\cite{mahmood2019amass} as testing data. 
% On both protocols, we adopt the SMPL~\cite{loper2015smpl}  model for human representation. 
Instead of using the same default shape for all motion sequences, which is widely used in previous work~\cite{aliakbarian2023hmdnemo, du2023avatars, jiang2022avatarposer, zheng2023realistic}, we utilize the ground-truth body shape parameters to calculate the joint positions.
 
\begin{figure*}[!tb]
    \centering
    \includegraphics[width=0.95\linewidth]{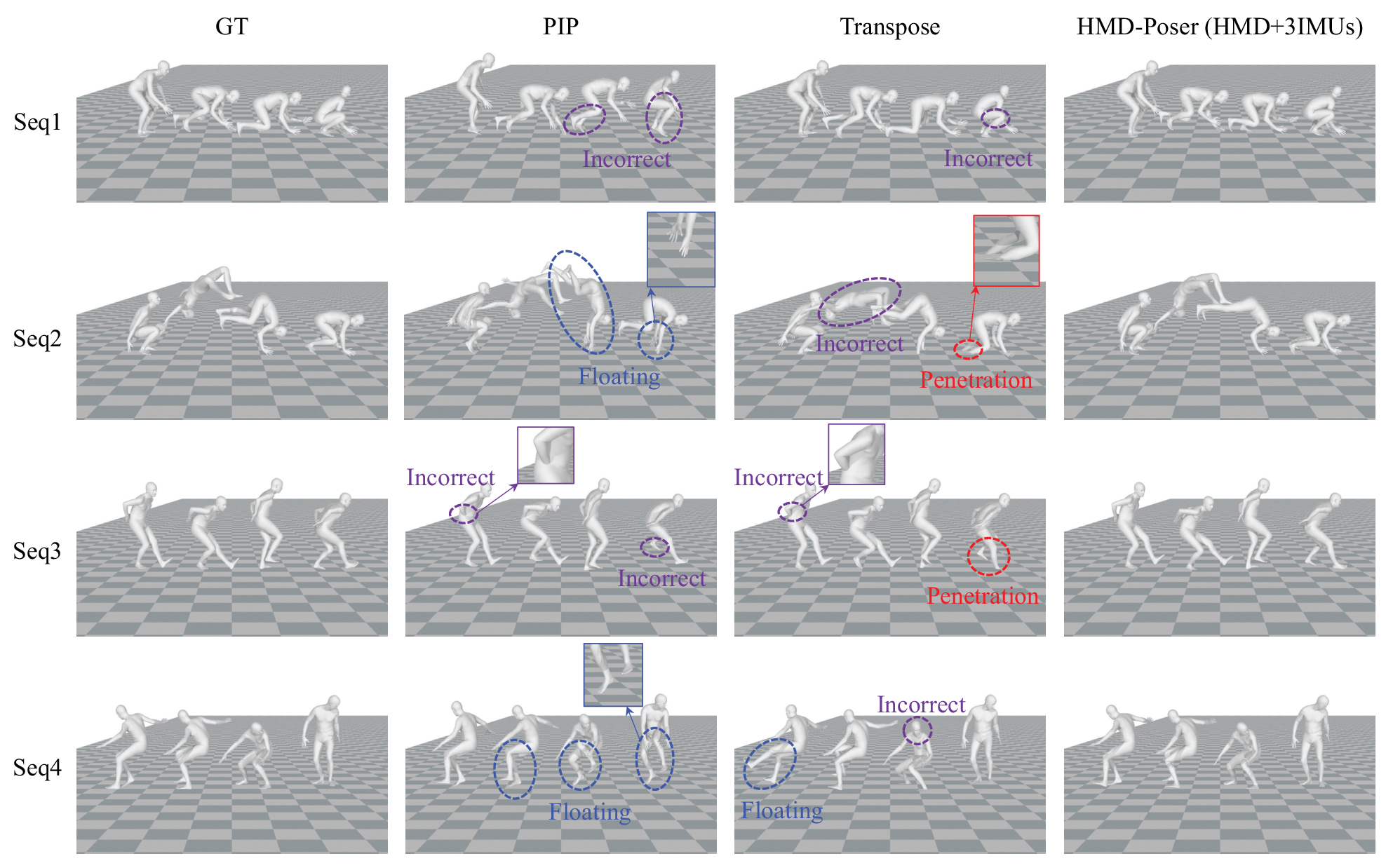}
    \caption{Qualitative comparisons between our method and 6IMUs-based methods. For a fair comparison, we provide head and hand positions to the baselines and compare them with our method under the HMD+3IMUs input scenario.}
    \label{fig:qualitative_6IMU}
\end{figure*}

\subsubsection{Comparison}
\label{sec:comparisontoSOTA}
% \noindent
% \textbf{Quantitative and Qualitative Results.}
As described in Sec.~\ref{sec:relatedwork}, there are two mainstream HMT methods, which use 6DOFs from HMD with hand controllers or use 3DOFs from IMUs respectively. We compare our HMD-Poser with state-of-the-art methods in both categories. Tab.~\ref{tab:results_protocol1} and Tab.~\ref{tab:results_protocol2} show the detailed quantitative results in the first and second protocols respectively. It can be concluded from these two tables that: (1) HMD-Poser surpasses all existing methods in both tracking accuracy (lower $\textit{MPJRE}$ and $\textit{MPJPE}$) and motion smoothness (lower $\textit{MPJVE}$ and $\textit{Jitter}$), introducing a new state-of-the-art on the AMASS dataset. (2) By adding IMU sensors to the lower legs and pelvis, all metrics especially the $\textit{L-PE}$ are significantly improved, proving the effectiveness of combining HMD with IMUs. (3) HMD-Poser in the HMD scenario, surpasses all HMD-based methods, i.e., AvatarPoser~\cite{jiang2022avatarposer}, AGRoL~\cite{du2023avatars} and AvatarJLM~\cite{zheng2023realistic}. It shows that our model can obtain the best results using the same data. We attribute this performance increase to our TSFL network generating rich temporal-spatial correlation information via combining LSTM with Transformer. We provide more comparison results in each scenario in the supplementary materials.

In Fig.~\ref{fig:qualitative_3IMU}, we show the comparison results between our HMD-Poser in the HMD scenario and previous HMD-based approaches. In Fig.~\ref{fig:qualitative_6IMU}, we also provide the comparison results between our HMD-Poser in the HMD+3IMUs scenario and previous works using six IMUs.  Obviously, our model could achieve better tracking accuracy and reduce abnormal issues such as floating and penetration.

\subsubsection{Ablation Studies}
\label{sec:ablationstudy}
%In this section, 
We ablate our method in various settings to validate the effect of the main components in HMD-Poser. 
% Without loss of generality, 
All ablation studies are conducted on protocol1 with the HMD scenario.

\noindent
\textbf{Effect of input representation.} 
As discussed in Sec.~\ref{sec:problemdefine}, HMD-Poser additionally introduces the hand representations relative to the head coordinate frame to improve joint tracking accuracy. Tab.~\ref{tab:handrep} presents the results with and without these representations. As shown, adding these input features could improve both the accuracy and smoothness of the generated motion. In particular, it reduces the $\textit{H-PE}$ by a large margin, which proves its effectiveness in estimating bone lengths on the chain from head to hands.

\begin{table}[ht]
  \centering
  \begin{tabular}{@{}p{2.9cm}cccc@{}}
    \toprule
    Method & MPJRE & MPJPE & H-PE & Jitter \\
    \midrule
    w/o $\{x^{t}_{lh/h}, x^{t}_{rh/h}\}$ &  2.45 & 3.43 & 2.36 & 6.25\\
    with $\{x^{t}_{lh/h}, x^{t}_{rh/h}\}$ & \textbf{2.28} & \textbf{3.19} & \textbf{1.65} & \textbf{6.07}\\
    \bottomrule
  \end{tabular}
  \caption{Evaluating the effect of adding hand representations relative to the head coordinate frame to input representation.}
  \label{tab:handrep}
\end{table}

\begin{table}[!htp]
  \centering
  \begin{tabular}{@{}p{2.9cm}cccc@{}}
    \toprule
    Method & MPJRE & MPJPE & H-PE & Jitter \\
    \midrule
    w/o ShapeHead &  2.32 & 5.08 & 4.25 & 6.11\\
    with ShapeHead & \textbf{2.28} & \textbf{3.19} & \textbf{1.65} & \textbf{6.07} \\
    \bottomrule
  \end{tabular}
  \caption{Evaluating the effect of the shape regression head. The default shape is used when there is no shape regression head.}
  \label{tab:abstushapehead}
\end{table}

\begin{table*}[!htp]
  \centering
  \begin{tabular}{@{}p{3.2cm}cccccccc@{}}
    \toprule
    Method & MPJRE$\downarrow$ & MPJPE$\downarrow$ & MPJVE$\downarrow$ & Jitter$\downarrow$ & H-PE$\downarrow$ & U-PE$\downarrow$ & L-PE$\downarrow$ & R-PE$\downarrow$ \\
    \midrule
    HMD-Poser (Online) & 6.48 & 6.55 & 30.60 & 16.96 & 8.10 & 5.25 & 8.52 & 7.13 \\
    HMD-Poser (Offline) & 6.45 & 6.53 & 30.56 & 16.95 & 8.01 & 5.20 & 8.46 & 6.98 \\
    HMD-Poser (Offline*) & \textbf{4.77} & \textbf{4.75} & \textbf{22.30} & \textbf{15.25} & \textbf{2.09} & \textbf{3.35} & \textbf{6.77} & \textbf{6.06} \\
    \bottomrule
  \end{tabular}
  \caption{Comparisons between our HMD-Poser method running offline and running on PICO 4 HMD. We choose the HMD+2IMUs input scenario for evaluation as it is suitable for most VR applications. * indicates results using synthetic input from ground-truth data.}
  \label{tab:resultsonpico}
  \vspace{-1em}
\end{table*}

% HMD-Poser can reach a frequency of 90.0Hz on PICO 4 HMD. To align with the setup in training, we set $\textit{FPS}$ to a fixed frequency of 60Hz for model inference.
 
\noindent
\textbf{Effect of the shape head.} 
% The shape regression head is another key module proposed in this paper.
Here, we compare the results with and without a shape regression head in Tab.~\ref{tab:abstushapehead}. It can be concluded that the shape regression head has a significant contribution to reducing position errors (lower $\textit{MPJPE}$ and $\textit{H-PE}$). This also indicates that joint position estimation is sensitive to differences in shape parameters among users, and it is unreasonable for previous methods to use the same default body shape for joint position calculation.

For more experiments on the effect of the model size and each loss term, please refer to our supplementary materials.

\subsection{Experiments on VR Devices}
\label{sec:liveperformance}
Existing human motion capture datasets, such as AMASS, are built on optical markers and do not contain HMD and IMU sensor data. So all existing methods used synthetic input signals from ground-truth data. And they did not cover some common issues in real VR applications, such as sensor measurement errors, calibration errors, etc. To investigate the model's performance gap between synthetic and real data and evaluate our HMD-Poser's performance running on HMDs, we built an additional dataset of real-captured data with HMD+2IMUs. It contains 74 free-dancing motions from 8 subjects (3 male and 5 female) wearing PICO 4 and 2 PICO motion trackers on the lower legs. Each motion sequence contains both the input HMD and IMU sensor data and the ground-truth SMPL parameters obtained via OptiTrack~\cite{optitrack} and Mosh++~\cite{loper2014mosh}. 
% To evaluate our HMD-Poser running on HMD devices, 
Meanwhile, a pre-trained model using AMASS data is deployed to PICO 4 HMD, and the model output is stored for evaluation. Please refer to supplementary materials for further details.

\subsubsection{Quantitative Results}
\label{sec:live_quantitative_res}
First, we make a quantitative comparison between our HMD-Poser method running offline and running on PICO 4 HMD. In this experiment, we all use real sensor data from HMD and IMUs. As shown in Tab.~\ref{tab:resultsonpico}, the performance gap between offline and online is small and it demonstrates that our HMD-Poser could run in real-time on portable HMDs with limited computing resources and achieve similar tracking performance as offline. Second, we compare the experimental results using synthetic and real data. The results using synthetic data are much better than those using real data, especially for the $\textit{H-PE}$ metric. The results are in line with expectations, as the connection between the user's hand and the hand controllers is not rigid all the time. When there are relative motions between them, the transformation matrix from the hand controller to the hand would deviate from the calibration results, resulting in large $\textit{H-PE}$.

\begin{figure}
    \centering
    \includegraphics[width=1.0\linewidth]{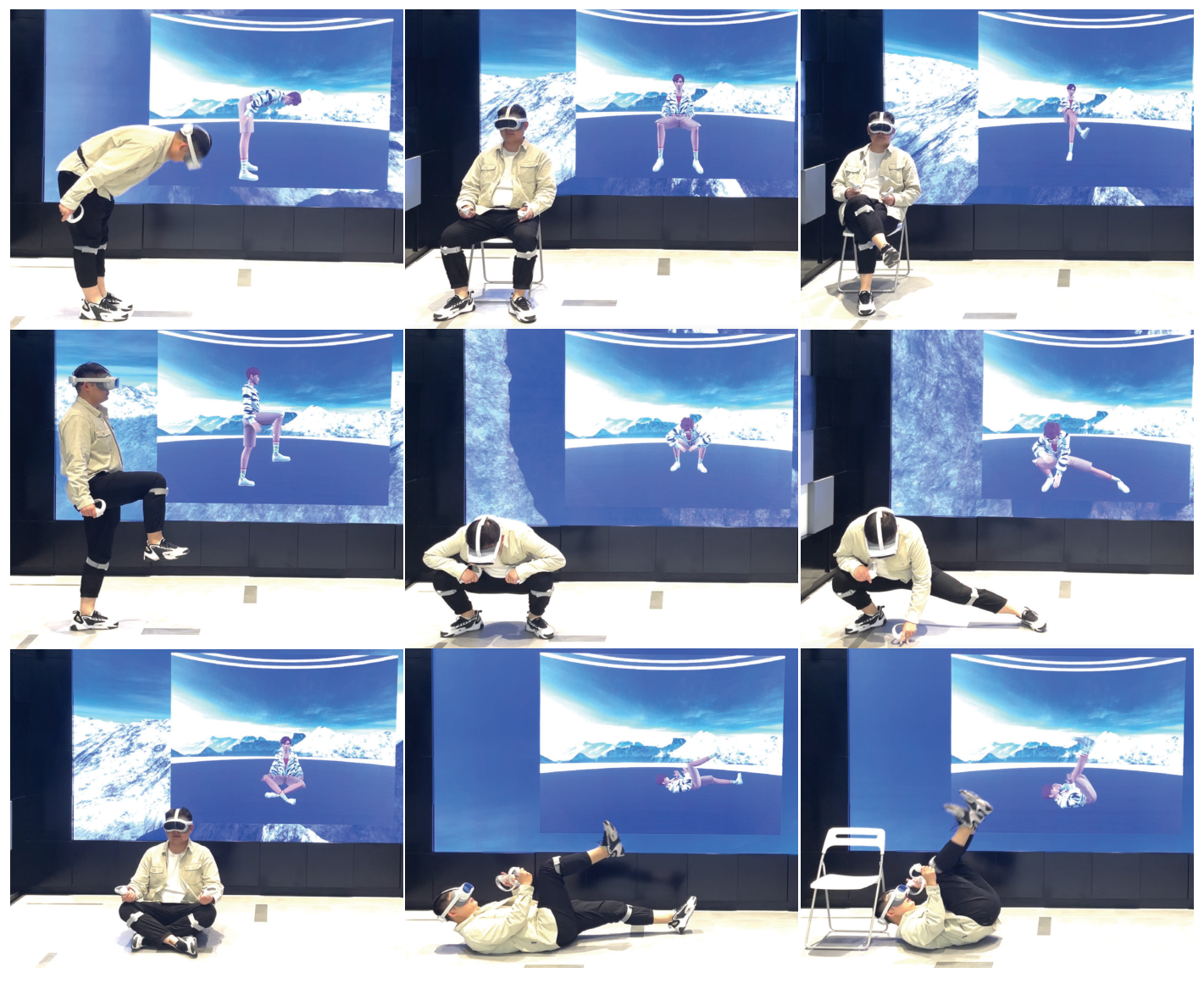}
    \caption{Results of real-time Avatar-driving on PICO 4 HMD.}
    \label{fig:quali_on_hmd}
    \vspace{-1em}
\end{figure}

\begin{table}[ht]
  \centering
  \begin{tabular}{@{}p{3.5cm}cc@{}}
    \toprule
    Method & FPS (GPU)$\uparrow$ & FPS (HMD)$\uparrow$ \\
    \midrule
    AvatarPoser~\cite{jiang2022avatarposer} &  114.1 & - \\
    AGRoL~\cite{du2023avatars} &  60.8 & - \\
    AvatarJLM~\cite{zheng2023realistic} &  1.9 & - \\
    Transpose~\cite{yi2021transpose} &  123.0 & - \\
    PIP~\cite{yi2022pip} &  62.5 & - \\
    HMD-Poser (Ours) & \textbf{205.7} & \textbf{90.0} \\
    \bottomrule
  \end{tabular}
  \caption{Comparison with baselines in terms of inference speed.}
  \label{tab:inference_speed}
  \vspace{-1em}
\end{table}

\subsubsection{Qualitative Results}
\label{sec:live_qualitative_res}
We qualitatively demonstrate our HMD-Poser's on-device performance with a real-time Avatar-driving application on PICO 4. As shown in Fig.~\ref{fig:quali_on_hmd}, the Avatar's motions are almost identical to those of the human. It demonstrates that our method could accurately reconstruct full-body motions in real-time in most sequences. Please refer to the videos in the supplementary materials for more qualitative results. 

\subsection{Inference Speed}
\label{sec:inference_speed}
For a fair comparison, we calculate the $\textit{FPS}$ of all methods running on the same NVIDIA GeForce RTX 3080 device, and present the results in Tab.~\ref{tab:inference_speed}. Owing to our lightweight TSFL network, HMD-Poser achieves an inference frequency of 205.7Hz on GPU and outperforms all existing methods by a large margin. Meanwhile, our HMD-Poser can reach a frequency of 90.0Hz on PICO 4 HMD, which has not been demonstrated in previous methods.

%%%%%%%%% Conclusion
% \section{Conclusion and Limitations}
\section{Conclusion}
\label{sec:conclusion}
In this paper, we present HMD-Poser, the first unified approach to recovering full-body motions with scalable sparse observations from HMD and wearable IMUs. Meanwhile, HMD-Poser proposes a lightweight TSFL network, making it deployable to consumer-level HMD devices and runnable in real-time. We experimentally demonstrate that our method achieves superior results with respect to state-of-the-art methods on the public AMASS dataset.In addition, we build an additional dataset of real-captured HMD and IMU data to validate that HMD-Poser could run online in portable HMDs and while maintaining similar tracking accuracy. We believe that our method paves a new way for human motion tracking on VR platforms.

\noindent
\textbf{Limitation and future works.} 
As a data-driven method, our approach is also highly dependent on large-scale training data, and more real-captured training data can also contribute greatly to the task.
Besides, due to the limitation of IMU itself, our method may struggle to disambiguate lower-body poses with similar measurements, such as slowly and uniformly lifting one foot vertically. 

{
    \small
    \bibliographystyle{ieeenat_fullname}
    \bibliography{main}
}


\end{document}

\title{HMD-Poser: On-Device Real-time Human Motion Tracking from Scalable Sparse Observations
\\(Supplementary Materials)}  % **** Enter the paper title here

\author{}

\renewcommand{\thesection}{\Alph{section}}  
\renewcommand{\thetable}{\Alph{table}}  
\renewcommand{\thefigure}{\Alph{figure}}

% \twocolumn[{%
% \renewcommand\twocolumn[1][]{#1}%
% \maketitle
% \thispagestyle{empty}

% % \begin{center}
% %     \vspace{-1.5em}
% %     % \vspace{-2em}
% %     \centering
% %     \captionsetup{type=figure}
% %     \includegraphics[width=0.95\textwidth ]{figure_3.eps}
% %         \captionof{figure}{Visual comparisons between various methods and ground truth.}
% %         \label{fig:visual comparisons with ap}
% % \end{center}%
% }]

\maketitle
\appendix

\section{Implementation Details}
% We present the details of each component of the HMD-Poser in this section.
\noindent
\textbf{Feature embedding.} This module comprises eight sub-modules to process eight components of the input data independently. Each sub-module consists of a few shallow multilayer perceptions (MLPs) to process rotation, angular velocity, position, linear velocity, and acceleration, respectively. Each MLP is a single $\rm Linear$ layer followed by $\rm LeakyReLU$. The outputs of the MLPs within each sub-module are concatenated to form a vector of size 256.

\noindent
\textbf{Lightweight TSFL.}
This module consists of two identical blocks, and each block has two sub-blocks: an LSTM-based block to learn temporal information and a Transformer-based block to encode spatial information. For each component of the input data, we use a separate unidirectional single-layer $\rm LSTM$ with a hidden size of 256 to encode the historical information.
The $\rm LSTM$ networks are orthogonally initialized.
To learn how different components are spatially correlated to each other, we adopt a 3-layer $\rm Transformer$ encoder with 8 attention heads and a feed-forward hidden size of 256 to process the outputs of LSTMs at each time step. The $\rm Transformer$ encoder itself can handle the scalable inputs via masks.

\noindent
\textbf{Regression heads.}
There are two regression heads that regress the local pose parameters and the shape parameters of SMPL~\cite{loper2015smpl} respectively. Both are MLP networks, and each MLP network consists of a $\rm Linear$ layer, $\rm LeakyReLU$ activation, and another $\rm Linear$ layer. We represent the joint orientations by the 6D reparametrization due to its simplicity and continuity~\cite{zhou2019continuity}. Therefore, the output feature dimension of the pose regression head is $22\times 6=132$. The output feature dimension of the shape regression head is set to 16. The intermediate dimension between the two $\rm Linear$ layers is set to 256.

\section{Additional Quantitative Comparisons}
As mentioned in the main paper, we consider three input scenarios in this paper, including (a) HMD, (b) HMD+2IMUs, and (c) HMD+3IMUs. In this section, we conduct extra experiments on each separate scenario. Meanwhile, to better demonstrate the superiority of our method, we also train a variant of the existing approaches, denoted as $\dagger$ in the following tables, by adding a shape regression head to their original model and introducing the joint position loss in the model training. All the experiments are also conducted on the AMASS dataset~\cite{mahmood2019amass} with two different protocols.
% The FPS of each method is calculated on the same NVIDIA GeForce RTX 3080 device.

\subsection{HMD Scenario}
In this scenario, we can fairly compare our method with state-of-the-art human motion tracking methods in the HMD setting, such as AvatarPoser~\cite{jiang2022avatarposer}, AGRoL~\cite{du2023avatars}, and AvatarJLM~\cite{zheng2023realistic}. We re-train these methods with their public source code and the ground truth body shape parameters. As shown in Tab.~\ref{tab:3IMUs}, HMD-Poser is not only more accurate (lower $\textit{MPJRE}$ and $\textit{MPJPE}$) but also generates smoother human motions (lower $\textit{Jitter}$) than all previous methods (without $\dagger$) on both protocol1 and protocol2. It validates that HMD-Poser achieves a new state-of-the-art on the AMASS dataset. By comparing HMD-Poser with the variants of previous methods, HMD-Poser has similar tracking accuracy to AvatarJLM$\dagger$~\cite{zheng2023realistic} on protocol2 but is significantly better than other methods on all protocols. Meanwhile, HMD-Poser is significantly better than AvatarJLM$\dagger$~\cite{zheng2023realistic} in inference speed, i.e., 205.7Hz vs 1.9Hz, as shown in the main paper.

\begin{table*}
  \centering
  \begin{tabular}{@{}l|l|ccccccccc@{}}
    \toprule
    Dataset & Method & MPJRE$\downarrow$ & MPJPE$\downarrow$ & MPJVE$\downarrow$ & Jitter$\downarrow$ & H-PE$\downarrow$ & U-PE$\downarrow$ & L-PE$\downarrow$ & R-PE$\downarrow$ \\
    \midrule
    \multirow{7}{*}{Protocol 1}
    & AvatarPoser~\cite{jiang2022avatarposer} &  2.94 & 5.84 & 26.60 & 13.97 & 4.58 & 3.24 & 9.59 & 5.05 \\
    & AvatarPoser$\dagger$~\cite{jiang2022avatarposer} &  2.94 & 5.17 & 27.17 & 14.45 & 3.64 & 2.71 & 8.73 & 4.39 \\
    & AGRoL~\cite{du2023avatars} &  2.70 & 5.73 & 19.08 & 7.65 & 4.29 & 3.16 & 9.44 & 5.15 \\ 
    & AGRoL$\dagger$~\cite{du2023avatars} &  3.32 & 6.58 & 23.81 & 11.45 & 4.32 & 3.38 & 11.20 & 5.77 \\
    & AvatarJLM~\cite{zheng2023realistic} &  2.81 & 5.03 & 20.91 & 6.94 & 2.01 & 3.00 & 7.96 & 4.58 \\
    & AvatarJLM$\dagger$~\cite{zheng2023realistic} &  2.56 & 3.89 & 20.91 & 7.46 & \textbf{1.60} & 2.01 & 6.62 & 3.32 \\
    & HMD-Poser(Ours) &  \textbf{2.28} & \textbf{3.19} & \textbf{17.47} & \textbf{6.07} & 1.65 & \textbf{1.67} & \textbf{5.40} & \textbf{3.02} \\
    \midrule
    \multirow{7}{*}{Protocol 2}
    & AvatarPoser~\cite{jiang2022avatarposer} & 4.68 & 6.62 & 33.16 & 10.79 & 3.93 & 2.97 & 11.89 & 5.30 \\
    & AvatarPoser$\dagger$~\cite{jiang2022avatarposer} & 4.64 & 6.63 & 33.54 & 10.77 & 3.30 & 2.81 & 12.14 & 5.42 \\
    & AGRoL~\cite{du2023avatars} & 4.38 & 6.74 & \textbf{24.14} & 6.33 & 3.53 & 3.02 & 12.11 & 5.86 \\
    & AGRoL$\dagger$~\cite{du2023avatars} & 4.82 & 8.17 & 33.82 & 15.75 & 5.69 & 3.75 & 14.56 & 6.63 \\
    & AvatarJLM~\cite{zheng2023realistic} & 4.45 & 5.96 & 27.50 & 6.91 & 2.30 & 2.97 & 10.28 & 5.22 \\
    & AvatarJLM$\dagger$~\cite{zheng2023realistic} & 4.28 & \textbf{5.43} & 27.14 & 6.89 & \textbf{1.88} & \textbf{2.32} & 9.93 & \textbf{4.67} \\
    & HMD-Poser(Ours) & \textbf{4.27} & 5.44 & 30.15 & \textbf{5.62} & 2.56 & 2.44 & \textbf{9.77} & 4.83 \\
    \bottomrule
  \end{tabular}
  \caption{Evaluation results in the HMD scenario. We retrain existing approaches with their public source code and the ground truth body shape parameters. $\dagger$ denotes a variation of the existing models by adding a shape regression head to their original model and introducing the joint position loss in the model training. The best results are in \textbf{bold}.}
  \label{tab:3IMUs}
\end{table*}

\begin{table*}
  \centering
  \begin{tabular}{@{}l|l|ccccccccc@{}}
    \toprule
    Dataset & Method & MPJRE$\downarrow$ & MPJPE$\downarrow$ & MPJVE$\downarrow$ & Jitter$\downarrow$ & H-PE$\downarrow$ & U-PE$\downarrow$ & L-PE$\downarrow$ & R-PE$\downarrow$ \\
    \midrule
    \multirow{7}{*}{Protocol 1}
    & AvatarPoser~\cite{jiang2022avatarposer} & 2.51 & 4.99 & 22.02 & 11.16 & 4.58 & 3.22 & 7.53 & 4.98\\
    & AvatarPoser$\dagger$~\cite{jiang2022avatarposer} & 2.52 & 4.24 & 23.17 & 12.09 & 3.69 & 2.68 & 6.49 & 4.23\\
    & AGRoL~\cite{du2023avatars} &  2.25 & 4.81 &15.13 & 8.44 & 4.25 & 3.09 & 7.28 & 4.95\\ 
    & AGRoL$\dagger$~\cite{du2023avatars} &  2.76 & 5.25 & 16.17 & 7.98 & 5.20 & 3.48 & 7.82 & 5.39\\
    & AvatarJLM~\cite{zheng2023realistic} &  2.38 & 4.24 & 18.72 & 7.39 & 2.00 & 2.90 & 6.16 & 4.34\\
    & AvatarJLM$\dagger$~\cite{zheng2023realistic} &  2.12 & 2.95 & 18.78 & 7.53 & 1.48 & 1.89 & 4.48 & 3.06\\
    & HMD-Poser(Ours) &  \textbf{1.83} & \textbf{2.27} & \textbf{13.28} & \textbf{5.96} & \textbf{1.39} & \textbf{1.51} & \textbf{3.35} & \textbf{2.74} \\
    \midrule
    \multirow{7}{*}{Protocol 2}
    & AvatarPoser~\cite{jiang2022avatarposer} & 3.87 & 4.58 & 25.98 & 9.75 & 3.74 & 2.88 & 7.03 & 4.99\\
    & AvatarPoser$\dagger$~\cite{jiang2022avatarposer} & 3.90 & 4.70 & 26.76 & 10.08 & 3.10 & 2.77 & 7.49 & 5.29 \\
    & AGRoL~\cite{du2023avatars} & \textbf{3.64} & 4.69 & \textbf{17.22} & 7.46 & 3.54 & 2.95 & 7.20 & 5.40 \\
    & AGRoL$\dagger$~\cite{du2023avatars} & 4.00 & 5.63 & 23.37 & 14.53 & 3.95 & 3.32 & 8.98 & 6.78 \\
    & AvatarJLM~\cite{zheng2023realistic} & 3.89 & 4.49 & 22.64 & 6.34 & 2.21 & 2.89 & 6.41 & 4.84 \\
    & AvatarJLM$\dagger$~\cite{zheng2023realistic} & 3.77 & 3.69 & 22.25 & 6.04 & 1.78 & 2.20 & 5.83 & 4.38 \\
    & HMD-Poser(Ours) & 3.66 & \textbf{3.68} & 20.29 & \textbf{6.22} & \textbf{1.65} & \textbf{2.14} & \textbf{5.92} & \textbf{4.51} \\
    \bottomrule
  \end{tabular}
  \caption{Evaluation results in the HMD+2IMUs scenario. Note that all previous methods are modified in this scenario by adding the IMU tracking signals to their input data and extending the dimension of their feature embedding layer.}
  \label{tab:5IMUs}
\end{table*}

% \begin{table}[!htp]
%   \centering
%   \begin{tabular}{@{}p{2.5cm}cccc@{}}
%     \toprule
%     Method & MPJRE & MPJPE & H-PE & Jitter \\
%     \midrule
%     $N=1$ & 2.40 & 3.31 & 1.74 & 11.77 \\
%     $N=2$ & \textbf{2.28} & 3.19 & 1.65 & 6.07 \\
%     $N=3$ & \textbf{2.28} & \textbf{3.18} & \textbf{1.60} & \textbf{5.20}\\
%     \bottomrule
%   \end{tabular}
%   \caption{Evaluating the effect of the number of blocks $N$ in temporal-spatial feature learning.}
%   \label{tab:abstubackbonesize}
% \end{table}

\subsection{HMD+2IMUs Scenario}
To the best of our knowledge, there is no available method for comparison in this scenario. We make a minor adjustment to the existing methods~\cite{du2023avatars, jiang2022avatarposer, zheng2023realistic} in the HMD setting by adding the IMU tracking signals to their input data and extending the dimension of their feature embedding layer. Following~\cite{yi2021transpose, yi2022pip}, we adopt synthesized IMU data on the AMASS dataset.
Detailed results are presented in Tab.~\ref{tab:5IMUs}.
Comparing the results in Tab.~\ref{tab:3IMUs} and Tab.~\ref{tab:5IMUs}, it can be concluded that the tracking accuracy, especially for the lower body, is significantly improved by adding the IMU signals from the lower legs. It demonstrates the effectiveness of our method by combining HMD with IMUs.
On protocol1, HMD-Poser surpasses all existing methods including their variants in all metrics.
On protocol2, HMD-Poser obtains the lowest position error and $\textit{Jitter}$ among all methods, but slightly higher $\textit{MPJRE}$ and $\textit{MPJVE}$ than AGRoL~\cite{du2023avatars}.

\subsection{HMD+3IMUs Scenario}
In this scenario, the input setting is closest to that of 6IMUs-based tracking methods. Therefore, we compare our HMD-Poser in the HMD+3IMUs scenario with state-of-the-art methods in this category, i.e., Transpose~\cite{yi2021transpose} and PIP~\cite{yi2022pip}. For a fair comparison, we add the global positions of the headset and hand controllers to the input data of the baselines~\cite{yi2021transpose, yi2022pip}.
The results are summarized in Tab.~\ref{tab:6IMUs}.
In this scenario, our HMD-Poser can surpass all previous methods in all metrics on both protocol1 and protocol2.
Comparing the results in Tab.~\ref{tab:5IMUs} and Tab.~\ref{tab:6IMUs}, the tracking accuracy of HMD-Poser is further improved which validates the effectiveness of adding IMU to the pelvis.

% \subsection{Computational complexity}
% As shown in Tab.~\ref{tab:3IMUs}, Tab.~\ref{tab:5IMUs}, and Tab.~\ref{tab:6IMUs}, our HMD-Poser outperforms all existing methods in terms of FPS owing to the lightweight TSFL network.
% Meanwhile, our HMD-Poser does not require any post-processing like other methods, such as inverse-kinematics module in AvatarPoser~\cite{jiang2022avatarposer} and physics-aware motion optimization in PIP~\cite{yi2022pip}.
% Such performance allows HMD-Poser to be deployed to HMDs.

\section{Additional Ablation Studies}

\begin{table*}[htp]
  \centering
  \begin{tabular}{@{}l|l| ccccccccc@{}}
    \toprule
    Dataset & Method & MPJRE$\downarrow$ & MPJPE$\downarrow$ & MPJVE$\downarrow$ & Jitter$\downarrow$ & H-PE$\downarrow$ & U-PE$\downarrow$ & L-PE$\downarrow$ & R-PE$\downarrow$ \\
    \midrule
    \multirow{5}{*}{Protocol 1}
    & Transpose~\cite{yi2021transpose} & 3.05 & 4.57 & 22.41 & 7.98 & 3.83 & 3.05 & 6.76 & 4.62 \\
    & TransPose$\dagger$~\cite{yi2021transpose} & 3.02 & 3.99 & 23.32 & 8.65 & 3.58 & 2.72 & 5.82 & 4.23 \\
    & PIP~\cite{yi2022pip} &  2.45 & 4.54 & 19.02 & 8.13 & 4.54 & 3.15 & 6.53 & 4.54 \\ 
    & PIP$\dagger$~\cite{yi2022pip} &  2.31 & 2.84 & 17.43& 6.99 & 3.00 & 2.16 & 3.82 & 2.86 \\
    & HMD-Poser(Ours) &  \textbf{1.73} & \textbf{1.89} & \textbf{11.03} & \textbf{5.35} & \textbf{1.27} & \textbf{1.46} & \textbf{2.46} & \textbf{2.37} \\
    \midrule
    \multirow{5}{*}{Protocol 2}
    & Transpose~\cite{yi2021transpose} & 4.31 & 5.29 & 28.18 & 5.16 & 7.38 & 3.86 & 7.36 & 4.80 \\
    & Transpose$\dagger$~\cite{yi2021transpose} & 3.94 & 4.73 & 29.11 & 6.02 & 5.60 & 3.42 & 6.61 & 4.57 \\
    & PIP~\cite{yi2022pip} & 3.61 & 4.16 & 22.22 & 6.89 & 4.28 & 2.97 & 5.89 & 4.30 \\
    & PIP$\dagger$~\cite{yi2022pip} & 3.80 & 4.21 & 26.55 & 7.54 & 4.97 & 3.04 & 5.90 & 4.28\\
    & HMD-Poser(Ours) & \textbf{3.49} & \textbf{3.13} & \textbf{16.17} & \textbf{4.93} & \textbf{1.81} & \textbf{2.17} & \textbf{4.51} & \textbf{3.88} \\
    \bottomrule
  \end{tabular}
  \caption{Evaluation results in the HMD+3IMUs scenario. For a fair comparison, we also add head and hand positions to the input data of all baseline methods.}
  \label{tab:6IMUs}
\end{table*}

\noindent
\textbf{Effect of the model size.}
The number of blocks $N$ in the lightweight TSFL network is a key hyper-parameter in our HMD-Poser. As shown in Tab.~\ref{tab:abstubackbonesize}, we see a clear downward tendency for both $\textit{MPJPE}$ and $\textit{Jitter}$ when increasing $N$ from 1 to 2. However, this tendency becomes less pronounced as $N$ continues to increase. Hence, we use $N=2$ in our final configuration which could obtain satisfactory results in both tracking accuracy and inference speed.

\noindent
\textbf{Effect of each loss term.}
As shown in the main paper, HMD-Poser is trained with five different loss terms.
Among these loss terms, $\mathcal{L}_{ori}$, $\mathcal{L}_{lrot}$ and $\mathcal{L}_{joint}$ are essential terms for model training.
We illustrate the contributions of the left two loss terms, i.e., $\mathcal{L}_{grot}$ and $\mathcal{L}_{smooth}$, in a leave-one-term-out manner.
As shown in Tab.~\ref{tab:abstuloss}, the smooth loss $\mathcal{L}_{smooth}$ has a positive impact on reducing the $\textit{MPJVE}$ and $\textit{Jitter}$ as expected. 
The global pose loss $\mathcal{L}_{grot}$ reduces the $\textit{MPJPE}$ and $\textit{H-PE}$, which may be attributed to its role in reducing the accumulating error of pose estimation along the kinematic chain.

\begin{figure}[tp]
    \centering
    \includegraphics[width=0.8\linewidth]{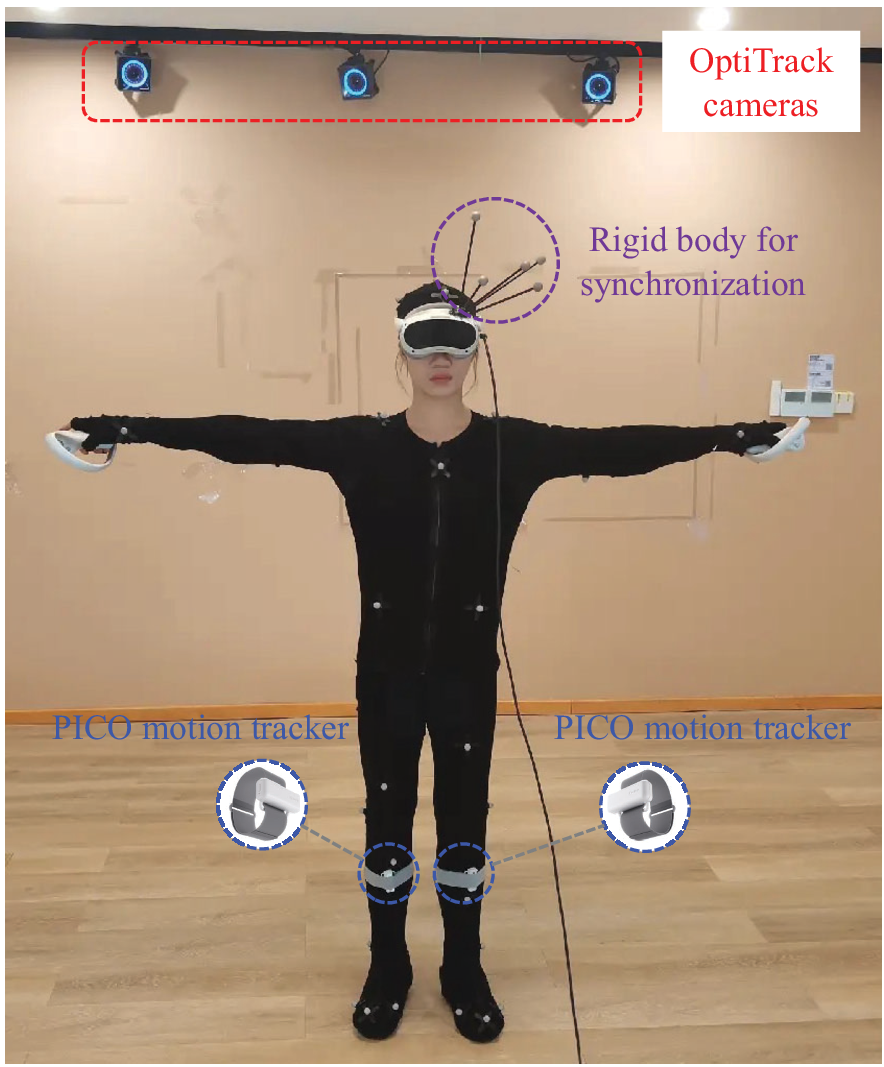}
    \caption{Setup for real-data collection with HMD + 2IMUs.}
    \label{fig:hmd_data_collection}
\end{figure}

\section{Real-Captured Data}
% We recorded 74 free-dancing motions from 8 subjects (3 male and 5 female) wearing PICO 4 (including HMD and two hand controllers) and 2 PICO motion trackers on the lower legs. The duration of each motion is set to 120 sec. Moreover, we use a synchronized marker-based motion capture system, OptiTrack~\cite{optitrack}, to track body markers and attain ground-truth SMPL parameters using Mosh++~\cite{loper2014mosh}. Hence, each motion sequence contains both the input HMD and IMU sensor data and the ground-truth SMPL parameters. To evaluate our HMD-Poser running on HMD devices, a pre-trained model is deployed to PICO 4 HMD, and the on-device $\textit{FPS}$ can reach a frequency of 90Hz. To align with the setup in AMASS, we set $\textit{FPS}$ to a fixed frequency of 60Hz. 
% All input and output data are stored for evaluation. Please refer to supplementary materials for further details.
To investigate the model's performance gap between synthetic data and real-captured sensor data and evaluate our HMD-Poser's performance running on HMDs, we built an additional dataset of real-captured data with HMD+2IMUs. As shown in Fig.~\ref{fig:hmd_data_collection}, each subject wears PICO 4 (including HMD and two hand controllers) and 2 PICO motion trackers on his/her lower legs and dances freely with music.
Meanwhile, we use a synchronized marker-based motion capture system, OptiTrack~\cite{optitrack}, to track body markers and attain ground-truth SMPL parameters using Mosh++~\cite{loper2014mosh}.
A total of 74 free-dancing motions from 8 subjects are recorded. The duration of each motion sequence is set to 120 seconds. We will release this dataset for research soon.
% We also provide a few demos of the generated motions in the supplementary materials.

\noindent
\textbf{Calibration.}
Since the raw IMU measurements are in the sensor-local coordinate system, we need to transform the raw IMU data into the same coordinate frame, which is referred to as calibration.
Although we have specified the rough wear positions of IMU sensors, i.e., the pelvis, and the left and right lower legs, there could still be differences in the precise wear position and orientation of each subject.
Our calibration method could automatically compute the transition matrices for each IMU sensor by requiring the subject to perform three specified actions: (1) stand straight for more than 5 seconds, (2) bend the knees forward and hold for 5 seconds, (3) lift the left and right legs in sequence. 
% We will explain the details of the calibration process in supplementary material.

\begin{table*}[!htp]
  \centering
  \begin{tabular}{@{}p{3cm}cccccccc@{}}
    \toprule
    Method & MPJRE$\downarrow$ & MPJPE$\downarrow$ & MPJVE$\downarrow$ & Jitter$\downarrow$ & H-PE$\downarrow$ & U-PE$\downarrow$ & L-PE$\downarrow$ & R-PE$\downarrow$ \\
    \midrule
    $N=1$ & 2.40 & 3.31 & 22.85 & 11.77 & 1.74 & 1.75 & 5.58 & 3.09 \\
    $N=2$ & \textbf{2.28} & 3.19 & 17.47 & 6.07 & 1.65 & 1.67 & 5.40 & \textbf{3.02} \\
    $N=3$ & \textbf{2.28} & \textbf{3.18} & \textbf{16.97} & \textbf{5.20} & \textbf{1.60} & \textbf{1.66} & \textbf{5.38} & 3.06 \\
    \bottomrule
  \end{tabular}
  \caption{Evaluating the effect of the number of blocks $N$ in the lightweight TSFL network.}
  \label{tab:abstubackbonesize}
\end{table*}

\begin{table*}[!htp]
  \centering
  \begin{tabular}{@{}p{3cm}cccccccc@{}}
    \toprule
    Method & MPJRE$\downarrow$ & MPJPE$\downarrow$ & MPJVE$\downarrow$ & Jitter$\downarrow$ & H-PE$\downarrow$ & U-PE$\downarrow$ & L-PE$\downarrow$ & R-PE$\downarrow$ \\
    \midrule
    w/o $\mathcal{L}_{grot}$ & 2.29 & 3.25 & 17.76 & \textbf{6.06} & 1.86 & 1.74 & 5.41 & 3.08 \\
    w/o $\mathcal{L}_{smooth}$ & \textbf{2.27} & \textbf{3.16} & 17.89 & 6.71 & 1.69 & 1.68 & \textbf{5.30} & \textbf{3.00} \\
    with all loss terms & 2.28 & 3.19 & \textbf{17.47} & 6.07 & \textbf{1.65} & \textbf{1.67} & 5.40 & 3.02 \\
    \bottomrule
  \end{tabular}
  \caption{Evaluating the effect of each loss term.}
  \label{tab:abstuloss}
\end{table*}

\noindent
\textbf{Synchronization.}
Since we jointly capture ground-truth motions and the sensor data of HMD and IMUs with separate devices, our records must be accurately synchronized in the absence of a genlock signal.
To this end, we add a rigid body on top of the HMD headset, as shown in Fig.~\ref{fig:hmd_data_collection}.
Subjects are asked to perform simple control movements at the beginning of each capture motion, consisting of turning their heads clockwise and counterclockwise, nodding, and shaking their heads.
This enables matching the orientations measured by IMUs on the HMD device with the rigid body orientations measured by OptiTrack.
The frame rates of OptiTrack (120Hz) and IMUs (500Hz) are sufficiently high and the synchronization error is negligible.

\noindent
\textbf{Downsampling.}
Our HMD-Poser can reach a frequency of 90.0Hz on PICO 4 HMD. To align with the setup in training, we set the $\textit{FPS}$ of our HMD-Poser to a fixed frequency of 60Hz on HMD devices. Hence, we also downsample the ground-truth motions from 120Hz to 60Hz.

{
    \small
    \bibliographystyle{ieeenat_fullname}
    \bibliography{main}
}